\documentclass{llncs}
\usepackage{float}
\usepackage{tabularx,ragged2e}
\usepackage{amsmath}
\usepackage{booktabs}
\usepackage{multirow}
\usepackage{subfigure}
\usepackage{lscape}
\usepackage[table,xcdraw]{xcolor}
\usepackage{makeidx}  
 \usepackage{multirow}
 \usepackage[table,xcdraw]{xcolor}
\usepackage{amssymb}
\usepackage{array}
\newcolumntype{P}[1]{>{\centering\arraybackslash}p{#1}}
\usepackage{graphicx}
\usepackage{tabu}
\usepackage[flushleft]{threeparttable}
\usepackage{url}
\usepackage{hyperref}
\makeindex
\begin{document}

\title{DFUC 2020: Analysis Towards Diabetic Foot Ulcer Detection}

\author{Bill Cassidy \inst{1} \and Neil D. Reeves \inst{2} \and Joseph M. Pappachan \inst{3} \and David Gillespie \inst{1} \and Claire O'Shea \inst{4} \and Satyan Rajbhandari \inst{3} \and Arun G. Maiya \inst{5} \and Eibe Frank \inst{6} \and Andrew J.M. Boulton \inst{7} \and David G. Armstrong \inst{8} \and Bijan Najafi \inst{9} \and Justina Wu \inst{4} \and Moi Hoon Yap \inst{1}\thanks{Corresponding author, email: M.Yap@mmu.ac.uk}}


\institute{Department of Computing and Mathematics, Faculty of Science and Engineering, Manchester Metropolitan University, Manchester, M1 5GD, UK \and Research Centre for Musculoskeletal Science \& Sports Medicine, Faculty of Science and Engineering, Manchester Metropolitan University, Manchester, M1 5GD, UK \and Lancashire Teaching Hospital, Preston PR2 9HT, UK \and Waikato District Health Board, New Zealand \and Manipal College of Health Professions, India \and Department of Computer Science, University of Waikato, New Zealand \and School of Medical Sciences, University of Manchester, UK \and Keck School of Medicine, University of Southern California, USA \and Baylor College of Medicine, Texas, USA}

\maketitle


\begin{abstract}
Every 20 seconds, a limb is amputated somewhere in the world due to diabetes. This is a global health problem that requires a global solution. The MICCAI challenge discussed in this paper, which concerns the automated detection of diabetic foot ulcers using machine learning techniques, will accelerate the development of innovative healthcare technology to address this unmet medical need. In an effort to improve patient care and reduce the strain on healthcare systems, recent research has focused on the creation of cloud-based detection algorithms. These can be consumed as a service by a mobile app that patients (or a carer, partner or family member) could use themselves at home to monitor their condition and to detect the appearance of a diabetic foot ulcer (DFU). Collaborative work between Manchester Metropolitan University, Lancashire Teaching Hospital and the Manchester University NHS Foundation Trust has created a repository of 4,000 DFU images for the purpose of supporting research toward more advanced methods of DFU detection. Based on a joint effort involving the lead scientists of the UK, US, India and New Zealand, this challenge will solicit original work, and promote interactions between researchers and interdisciplinary collaborations. This paper presents a dataset description and analysis, assessment methods, benchmark algorithms and initial evaluation results. It facilitates the challenge by providing useful insights into state-of-the-art and ongoing research. This grand challenge takes on even greater urgency in a peri and post-pandemic period, where stresses on resource utilization will increase the need for technology that allows people to remain active, healthy and intact in their home.
\end{abstract}

\section{Introduction}
Wounds on the feet known as Diabetic Foot Ulcers (DFUs) are a major complication of diabetes. DFUs can become infected, leading to amputation of the foot or lower limb. Patients who undergo amputation experience significantly reduced survival rates \cite{soo2020diabetic}. In previous studies, various researchers \cite{van2013infrared}\cite{wang2015unified}\cite{goyal2017fully}\cite{goyal2018dfunet}\cite{goyal2020recognition} have achieved high accuracy in the recognition of DFUs using machine learning algorithms. Additionally, researchers have demonstrated proof-of-concept in studies using mobile devices for foot image capture \cite{yap2018new} and DFU detection \cite{goyal2018robust}. However, there are still gaps in implementing these technologies across multiple devices and locations in real-world settings. To bridge these gaps, we bring world-leading researchers from international institutions to work collaboratively towards automatic DFU detection.

The goal of the Diabetic Foot Ulcers Grand Challenge 2020 (DFUC 2020) \cite{yap2020dfuc} is to improve the accuracy of DFU detection in a real-world setting, and to motivate the use of more advanced machine learning techniques that are data-driven in nature. In turn, this will aid the development of a mobile app that can be used by patients, their carers, or their family members to help with remote detection and monitoring of DFU in a home setting. Enabling patients to engage in active surveillance outside of the hospital will reduce risk for the patient and commensurately reduce resource utilization by health care systems \cite{rogers2020covid}\cite{rogers2020wound}. This is particularly pertinent in the current post-COVID-19 climate. People with diabetes have been shown to be at higher risk of serious complications from COVID-19 infection \cite{ada2020covid19}, therefore limiting exposure is a priority.

\section{Related Work}
Recent years have attracted a growth in research interest in DFU due to the significantly increased number of reported cases of diabetes and the growing burden this represents on healthcare systems. Goyal et al. \cite{goyal2018robust} trained and validated a supervised deep learning model capable of DFU localisation. The backbone used was Faster R-CNN with Inception V2. This single classifier model was trained using two-tier (partial and full) transfer learning with the MS COCO dataset. Their method is capable of multiple detections per image, and demonstrated high mAP in experimental settings. However, this experiment used a relatively small dataset of 1,775 DFU images, with a post-processing stage required to remove false positives. Hence, the study is inconclusive for practical use of the proposed method in real-world settings. Additionally, this study was conducted in 2018. Improved object detection methods have emerged since that time, such as the very recently proposed EfficientDet \cite{tan2019effdet}. These newer methods may provide superior accuracy, smaller model sizes and improved inference times.

Wang et al. \cite{wang2016area} created a mirror image capture box to aid the process of obtaining DFU photographs for serial analysis. This study implemented a cascaded two-stage Support Vector Machine classification to determine DFU area. Segmentation and feature extraction was achieved using a superpixel technique to perform two-stage classification. One of these experiments included the use of a mobile app with the capture box \cite{wang2014smartphone}. Although the solution is highly novel, the system exhibited a number of limitations. The mobile app solution is constrained by the processing power available on the mobile device. Deep learning models can be adapted for mobile use, but usually at the expense of inference accuracy. The analysis requires physical contact between the capture box and the patient’s foot. This raises concerns of microbial contamination, especially in a medical setting where the box would be used by multiple patients. If the patient has not been diagnosed with peripheral neuropathy, and therefore still has sensation in their feet, contact with the capture box may introduce discomfort or pain. Marks or residue on the glass surface could also influence the image analysis. Wound contact with the glass surface could distort the size or shape of the wound between capture sessions, which could complicate serial monitoring. The design of the capture box also limits monitoring of DFU to those that appear on the plantar surface of the foot. Additionally, the sample size of the experiment was small, with only 35 images from real patients, and 30 images of moulage wound simulation. A more substantial sample size would need to be analysed to determine system effectiveness.

Brown et al. \cite{brown2017myfootcare} created a mobile app called MyFootCare, which attempts to promote patient self-care using personal goals, diaries, and notifications. The app maintains a serial photographic record of the patient’s feet. DFU segmentation is completed using a semi-automated process, where the user manually delineates the DFU location and surrounding skin tissue. The development of the app was informed by Fogg's behaviour model for persuasive technology, which stipulates that people need the ability, motivation, and triggers to enact desirable behaviour. MyFootCare allows patients to take photographs of their feet by placing the phone on the floor. The patient places their foot above the phone screen, and the photograph is automatically taken when the foot is correctly positioned. Voice feedback is used to guide the user when positioning their foot. However, this feature was not used during the experiment reported in \cite{brown2017myfootcare}, so its efficacy is unknown at this stage.

\section{Methodology}
This section discusses the DFU dataset, its ground truth labelling and baseline approaches to benchmark the performance of detections, submission rules and assessment methods.

\subsection{Dataset and Ground Truth}
We have received approval from the UK National Health Service (NHS) Research Ethics Committee (REC) to use these images for the purpose of research. The NHS REC reference number is 15/NW/0539. Foot images displaying DFU were collected from Lancashire Teaching Hospital (LTH) over the past few years. Three cameras were used for capturing the foot images: Kodak DX4530, Nikon D3300 and Nikon COOLPIX P100. The images were acquired with close-ups of the foot at a distance of around 30–40 cm with the parallel orientation to the plane of an ulcer. The use of flash as the primary light source was avoided, and instead, adequate room lights were used to ensure consistent colors in the resulting photographs. The images were acquired by medical photographers with specialization in the diabetic foot. As a preprocessing stage, we have discarded photographs that were excessively out of focus. We also excluded duplicates, identified by hash value for each file. 

\begin{figure}
	\centering
	\includegraphics[scale=0.35]{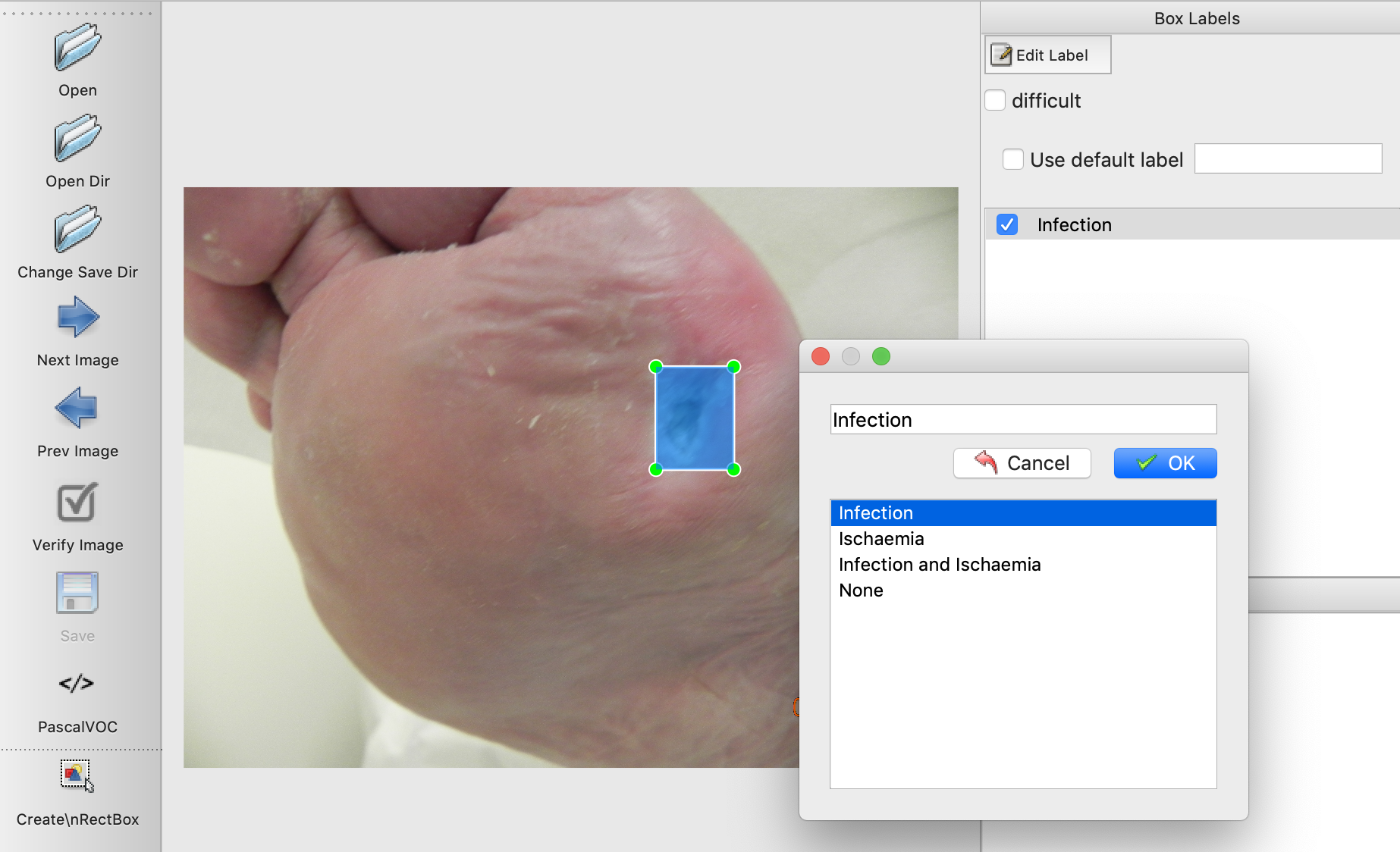}
	\caption{Experts' annotation of the region of interest and the pathology label. Courtesy of LabelImg \cite{tzutalingit}.}
	\label{fig:overview}
\end{figure}

The dataset for DFUC 2020 consists of 4,000 images, with 2,000 used for the training set and 2,000 used for the testing set. Additionally, 200 images were used for a sanity check on the grand challenge website. The training set consists of DFU images only, and the testing set comprised of images of DFU, other foot/skin conditions and images of healthy feet. The dataset is heterogeneous, with aspects such as distance, angle, orientation, lighting, focus, and the presence of background objects all varying between photographs. We consider this element of the dataset to be important, given that future models will need to account for numerous environmental factors in a system being used in non-medical settings. The images were captured during regular patient appointments at the LTH foot clinic, therefore some images were taken from the same subjects at different intervals. This means that the same ulcer may be present in the dataset more than once, but at different stages of development, at different angles and lighting conditions.

The following describes other notable elements of the dataset, where a case refers to a single image:

\begin{itemize}
    \item Cases may exhibit more than one DFU.
    \item Cases exhibit DFU at different healing stages.
    \item Cases may not always show all of the foot.
	\item Cases may show one or two feet, although there may not always be a DFU on each foot.
	\item Cases may exhibit partial amputations of the foot.
	\item Cases may exhibit deformity of the foot of varying degrees, a result of Charcot arthropathy.
	\item Cases may exhibit background objects, such as medical equipment, Doctor's hands, or wound dressings.
	\item Cases may exhibit partial blurring.
	\item Cases may exhibit partial obfuscation of the wound by medical instruments.
	\item Cases may exhibit signs of debridement, the area of which is often much larger than the ulcer itself.
	\item Cases may exhibit the presence of all or part of a toenail within a bounding box.
	\item Cases may exhibit subjects of a variety of ethnicities, although the majority are of white ethnicity.
	\item Cases may exhibit signs of infection and / or ischemia.
	\item Cases may exhibit the patient's face. In these instances, the face has been blurred.
	\item Cases may exhibit a time stamp printed on the image. If a DFU is obfuscated by a time stamp, the bounding box was adjusted to include as much of the wound as possible, while excluding the time stamp.
	\item Cases may exhibit imprint patterns resulting from close contact with wound dressing materials.
	\item Cases may exhibit unmarked circular stickers or rulers placed close to the wound area, used as a reference point for wound size measurement. Bounding boxes were adjusted to exclude rulers.
\end{itemize}

All training, validation and test cases are annotated with the location of foot ulcers in \textit{xmin}, \textit{ymin}, \textit{xmax} and \textit{ymax} coordinates, as illustrated in Figure 1. Two annotation tools were used to annotate the images - LabelImg \cite{tzutalingit} and VGG Image Annotator \cite{dutta2016via}. These were used to annotate images with a bounding box which indicates the ulcer location. The ground truth was produced by two healthcare professionals who specialize in treating and managing diabetic foot ulcers and associated pathology (a podiatrist and a consultant physician with specialization in the diabetic foot, both with more than 5 years professional experience). The instruction for annotation was to label each ulcer with a bounding box. If there was disagreement on DFU annotations, the final decision was mutually settled with the consent of both.

\begin{figure}
	\centering
	\begin{tabular}{ccc}
		\includegraphics[width=4cm,height=3cm]{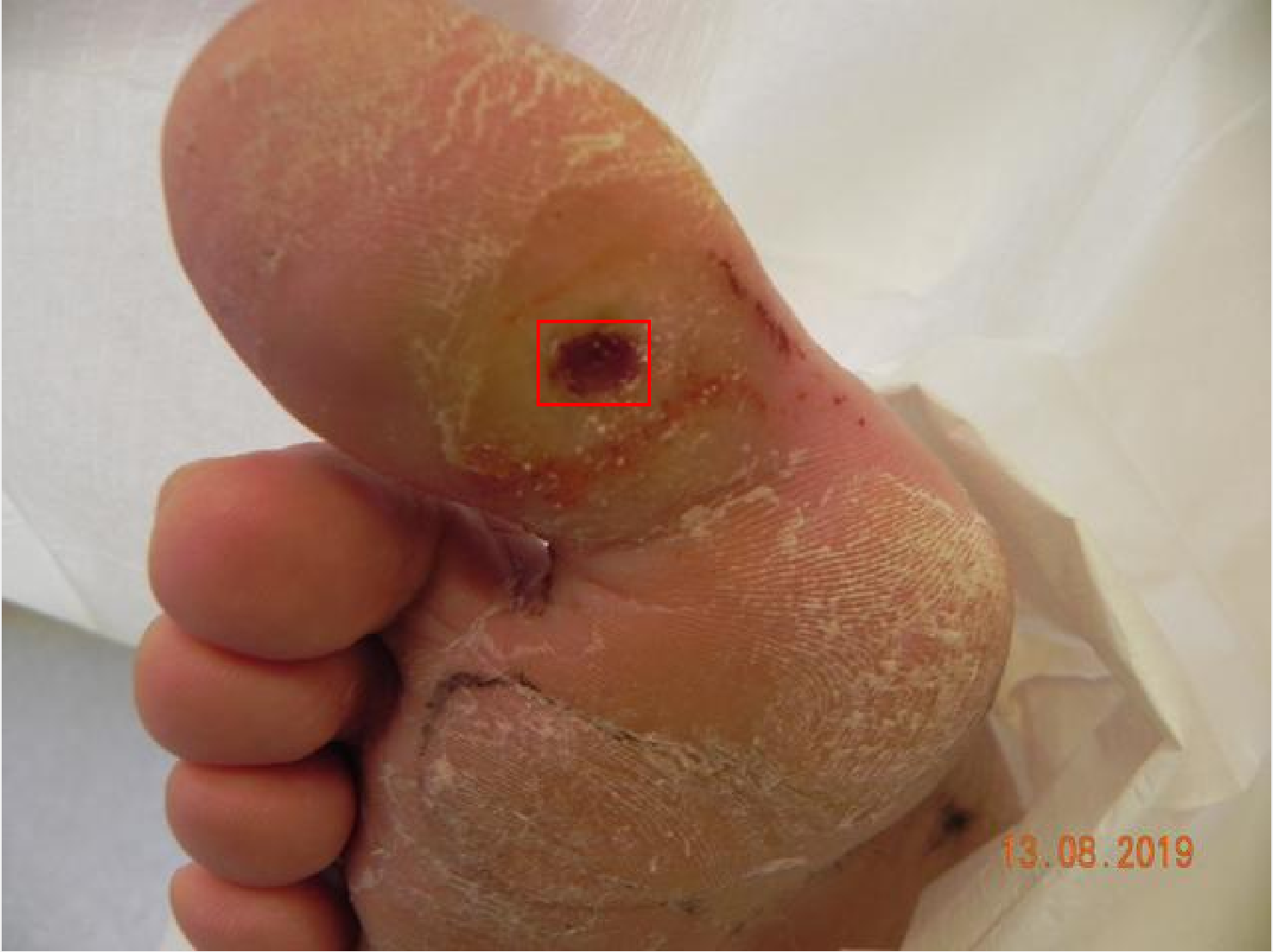} &
		\includegraphics[width=2.5cm,height=2.5cm]{resize.png} &
		\includegraphics[width=3cm,height=2cm]{resize.png}\\
		(a) & (b) & (c)\\  
	\end{tabular}     
	\caption[]{Illustration of the image resizing methods: (a) Original image; (b) Image resized by Goyal et al. \cite{goyal2018robust}; and (c) Image resized by our method.}
	\label{fig:resize}
\end{figure}

In this dataset, the size of foot images varies between 1600 $\times$ 1200 and 3648 $\times$ 2736 pixels. For the release dataset, we resized all images to 640 $\times$ 480 pixels to reduce computational costs during training. Unlike the approach by Goyal et al. \cite{goyal2018robust}, we preserve the aspect ratio of the images using the high quality anti-alias downsampling filter method found in the Python Imaging Library \cite{lund2020pillow}. Figure \ref{fig:resize}(a) shows the original image with the ground truth annotation. Figure \ref{fig:resize}(b) shows the resized image by Goyal et al. \cite{goyal2018robust} where the ulcer size and shape changed. We keep the aspect ratio while resizing, as illustrated in Figure \ref{fig:resize}.

\begin{figure}
	\centering
	\includegraphics[scale=0.33]{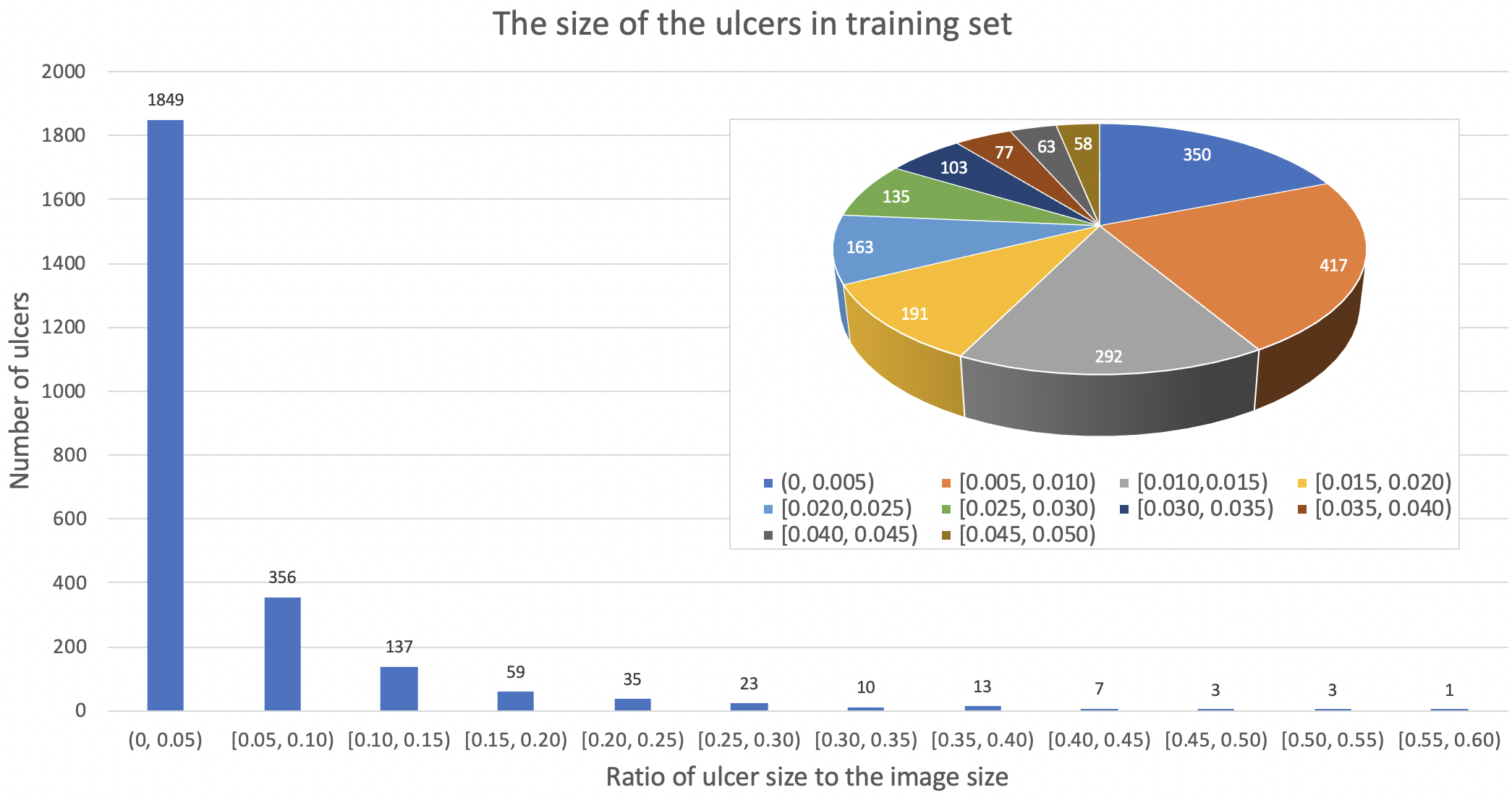}
	\caption{The ratio of annotated DFUs to foot images in the training set.}
	\label{fig:trainingset}
\end{figure}

For the training set, there are a total of 2,496 ulcers. A number of images exhibited more than one foot, or more than one ulcer, hence the discrepancy between the number of images and the number of ulcers. The size distribution of the ulcers in proportion to the foot image size is presented in Figure \ref{fig:trainingset}. We observed that the size for the majority of ulcers (1849 images, 74.08\%) is less than 5\% of the image size, in most cases indicating that the size of ulcers is relatively small. When conducting further analysis on these images (as illustrated by the pie chart in Figure \ref{fig:trainingset}), we found the majority of ulcers (1,250 images, 50.08\%) are less than 2\% of the image size.

\subsection{Assessment Methods}
To enable a fair technical comparison in the challenge, participants were not permitted to use an external training dataset unless they agreed that it could be shared with the research community. Participants were also encouraged to report the effect of using a larger training dataset on their techniques. This section discusses the metrics and ranking methods for DFUC 2020.

\subsubsection{Performance Metrics}
The F1-score and Mean average precision (mAP) will be used to assess the performance of the proposed algorithms. Participants were required to record all their detections (including multiple object detections) in a log file. A true positive is obtained when the Intersection over Union (IoU) of the bounding box is greater or equal to 0.5, which is defined by:
\begin{equation}
\frac{BB_{detected}\cap BB_{groundTruth}}{BB_{detected}\cup BB_{groundTruth}} \geq 0.5
\end{equation}
where $BB_{groundTruth}$is the bounding box provided by the experts on ulcer location, and $BB_{detected}$ is the bounding box detected by the algorithm.

F1-Score is the harmonic mean of Precision and Recall and provides a more suitable measure of predictive performance than the plain percentage of correct predictions in this application. F1-score is used as the False Negatives and False Positives are crucial, while the number of True Negatives can be considered less important. False Positives will result in additional cost and time burden to foot clinics, while False Negatives will risk further foot complications. The relevant mathematical expressions are as follows:
\begin{equation}
Recall = \frac{TP}{TP+FN}
\end{equation}

\begin{equation}
Precision = \frac{TP}{TP+FP}
\end{equation}

\begin{equation}
F1-score = 2 \times \frac{Recall \times Precision}{Recall+Precision}=\frac{2TP}{2TP+FP+FN}
\end{equation}

where TP is the total number of True Positives, FP is the total number of False Positives and FN is the total number of False Negatives.

In the field of object detection, a more widely accepted performance metric is mAP. This metric is used extensively to measure the overlap percentage of the prediction and ground truth [2], and is defined as the average of Average Precision for all classes:

\begin{equation}
mAP=\frac{{\sum^{Q} _{q=1}} AveP(q) }{Q}
\end{equation}

where Q is the number of queries in the set, and AveP(q) is the average precision for a given query, q. The Area Under the Precision Recall Curve (AUCPR) is interpolated to reduce the small variations in the ranking of detections. The exact method of mAP calculation can vary between networks and datasets, often depending on the size of the object that the network has been trained to identify.

\subsubsection{Ranking Methods}
Participants were ranked according to F1-score and mAP. However, for the completeness of scientific assessment, other metrics (precision and recall) are also be reported. All missing results, i.e. images with no labelled coordinates, are treated as no DFU detected on the image. If there are ties in both F1-score and mAP, submission time stamps were used for determining rank, ensuring that participants who provided the best solution in the shortest time were rewarded.
 
\section{Benchmark Algorithms}
To benchmark the dataset, we conduct experiments with three popular deep learning object detection networks: Faster R-CNN \cite{ren2015faster}, YOLOv5 \cite{jocher2020yolov5} and EfficientDet \cite{tan2019effdet,xuannianz2020edkeras}. Each of these networks is described in the following sub-sections.

\subsection{Faster R-CNN}
Faster R-CNN was introduced by Ren et al. \cite{ren2015faster}. This network is comprised of three sub-networks - a feature network, a region proposal network (RPN) and detection network (R-CNN). The feature network extracts features from an image that are then passed to the RPN which generates a series of proposals. These proposals represent locations where objects (of any type) have been initially detected (regions of interest). The outputs from both the feature network and the RPN is then passed to the detection network, which further refines the RPN output and generates the bounding boxes for the objects that have been detected. Non-Maximum Suppression and bounding box regression are used to eliminate duplicate detections and to optimize the box position around the detected object \cite{goswami2018frcnn}.

\subsection{YOLOv5}
YOLO (You Only Look Once) was introduced by Redmon et al. \cite{redmon2016you}. The authors focus on speed and aim for real-time object detection. Since then, it has become widely used in object detection with the latest versions being YOLOv4 \cite{bochkovskiy2020yolov4} and YOLOv5, produced by other authors. YOLOv5 requires an image to be passed through the network only once. A data loader is used for automatic data augmentation which has three stages: (1) scaling, (2) color space adjustment, and (3) mosaic augmentation. The mosaic augmentation combines four images into four tiles of random ratios. This is used as a means to overcome the limitation of older YOLO networks ability to detect smaller objects. A single CNN is used to process multiple predictions and class probabilities. Non-max suppression is used to ensure that each object in an image is only detected once \cite{odsc2018yolo}.

\subsection{EfficientDet}
EfficientNet (classification network) and EfficientDet (object detection network) were introduced by Tan et al. \cite{tan2019effnet,tan2019effdet}. EfficientDet applies feature fusion (a bidirectional feature pyramid network (BiFPN)) to input images to combine representations of an image at different resolutions. The last few outputs from the network backbone (EfficientNet) are used for the feature fusion stage. Learnable weights are applied at this stage so the network can determine which combinations from the BiFPN contribute to the most confident predictions. The final stage uses the BiFPN outputs to predict class and to plot the positions of bounding boxes. One of the most prominent features of EfficientDet is its scalability, allowing all three sub-networks (and image resolution) to be jointly scaled. This allows for the network to be tuned for different target hardware platforms to accommodate differences in hardware capability \cite{solawetz2020effdet,tan2020effdet}.

\subsection{Benchmark Experiments}
\begin{figure}
	\centering
	\includegraphics[scale=0.75]{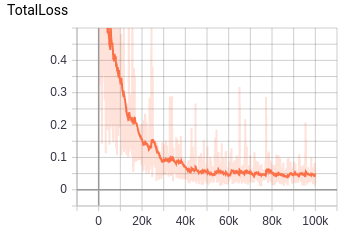}
	\caption{Training loss for Inception-v2-ResNet101.}
	\label{fig:trainingloss}
\end{figure}

For Faster R-CNN, we assess the performance of three different backbones: ResNet101, Inception-v2-ResNet101 and R-FCN. For the experimental settings, we used a batch size of 2 and ran gradient descent for 100 epochs initially to observe the loss. We began with a learning rate of 0.002, then reduced this to 0.0002 in epoch 40, and subsequently to 0.00002 in epoch 60. Figure \ref{fig:trainingloss} shows the training loss of the dataset for Inception-v2-ResNet101. It is noted that the network started to converge at 60 epochs. Hence, for Faster R-CNN, we train the models for 60 epochs.

For the EfficientDet experiment settings, training was completed using Adam stochastic optimisation, with a batch size of 32 for 50 epochs (1000 steps per epoch) and a learning rate of 0.001. For transfer learning, the EfficientNet-B0 backbone was used during training. Random transforms were used as a pre-processing stage while training the EfficientDet network to provide automatic data augmentation.

To benchmark the performance of the YOLO network on our dataset, we implement YOLOv5. This is due to the simplicity in installation and superior training and inference times compared to older iterations. For our implementation, we use a batch size of 8, and a pre-trained model from MS COCO `yolov5s' provided by the originator of YOLOv5 \cite{jocher2020yolov5}. We use the default settings for all the parameters in this experiment.


The system configuration used for the R-FCN, Faster R-CNN ResNet-101, Faster R-CNN Inception-v2-ResNet101 and YOLOv5 experiments were: (1) Hardware: CPU - Intel i7-6700 @ 4.00Ghz, GPU - NVIDIA TITAN X 12Gb, RAM - 32GB DDR4 (2) Software: Ubuntu Linux 16.04 and Tensorflow. The system configuration used for the EfficientDet experiment was: (1) Hardware: CPU - Intel i7-8700 @ 4.6Ghz, GPU - EVGA GTX 1080 Ti SC 11GB GDDR5X, RAM - 16GB DDR4 (2) Software: Ubuntu Linux 20.04 LTS with Keras and Tensorflow.

\begin{figure}
	\centering
	\begin{tabular}{cc}
		\includegraphics[width=6cm,height=4.5cm]{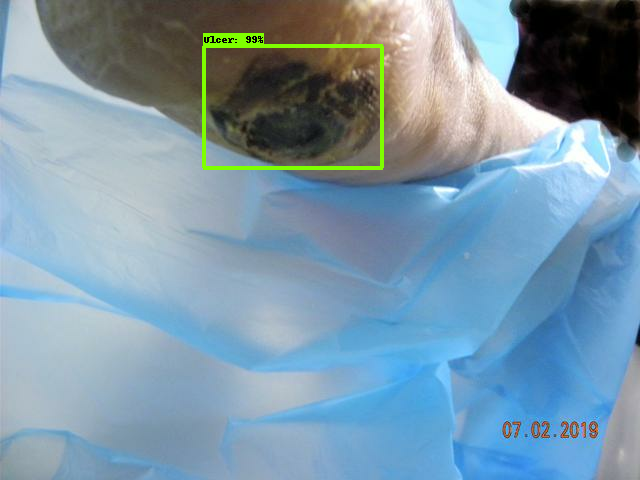} &
		\includegraphics[width=6cm,height=4.5cm]{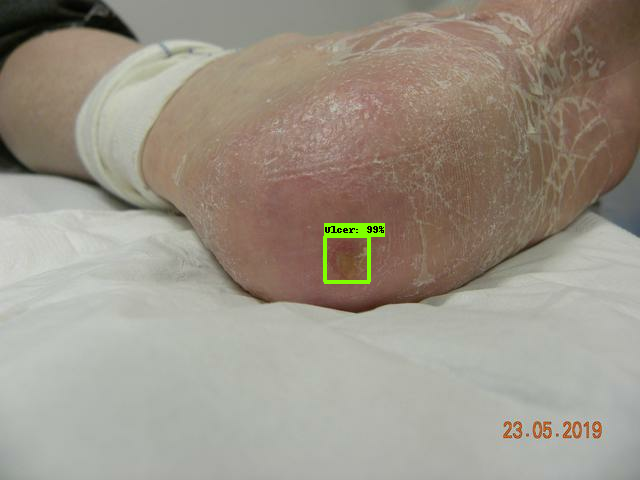}\\
		(a) &(b) \\  
	\end{tabular}     
	
	\caption[]{Illustration of the single detection result.}
	\label{fig:result}
\end{figure}

\begin{figure}
	\centering
	\begin{tabular}{cc}
		\includegraphics[width=6cm,height=4.5cm]{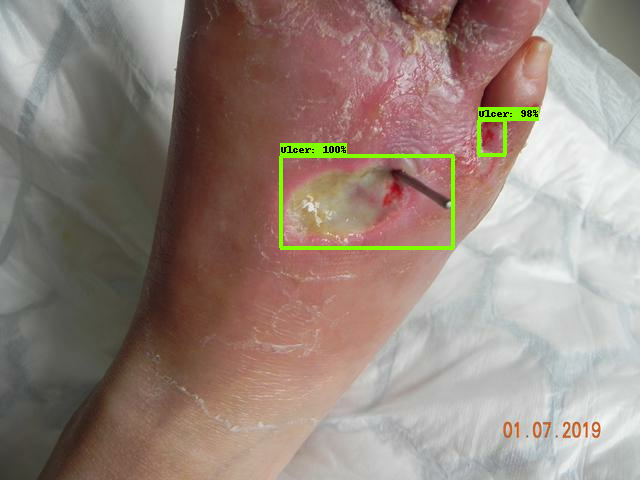} &
		\includegraphics[width=6cm,height=4.5cm]{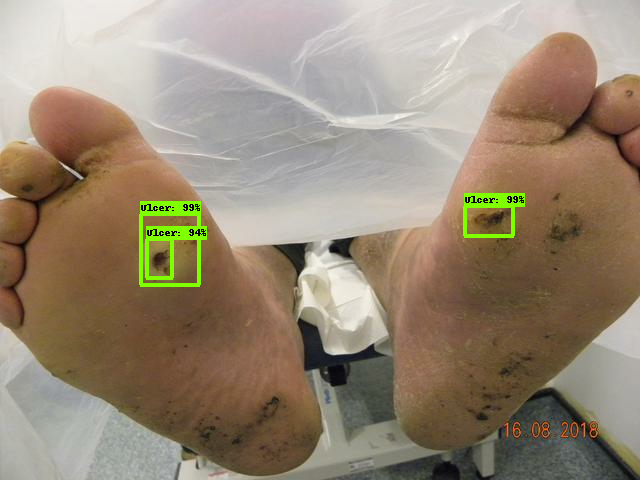} \\
		(a) &(b) \\  
	\end{tabular}     
	
	\caption[]{Illustration of multiple detection results.}
	\label{fig:result_multiple}
\end{figure}

The trained detection models detected single regions with high confidence as illustrated in Figure \ref{fig:result}. Additionally, each trained model detected multiple regions as illustrated in Figure \ref{fig:result_multiple}. At this stage, no post-processing was implemented. 

\section{Results and Analysis}
We evaluate the performance of the baseline algorithms without post-processing stages. First we compare the precision, recall, F1-Score and mAP of the baseline algorithms at IoU$\geq$0.5, then we compare the mAP at IoU of 0.5 to 0.90, with an increment of 0.10. 

\begin{table*}[!ht]
  \centering
  \small\addtolength{\tabcolsep}{2pt}
  \renewcommand{\arraystretch}{1.5}
  \caption{Performance of the benchmark algorithms on the testing set. FRCNN represents Faster R-CNN.}
  \label{table:Results}
  \scalebox{1.0}{
    \begin{tabular}{ccccc}
      \hline
        Benchmark Algorithm  & Recall    & Precision  & F1-Score   & mAP            \\ \hline\hline
        FRCNN R-FCN   & 0.7511     & 0.6186   & 0.6784    &\textbf{0.6596} \\
        FRCNN ResNet101  & 0.7396 & 0.5995    & 0.6623    & 0.6518         \\
        FRCNN Inception-v2-ResNet101 &\textbf{0.7554}  & 0.6046  & 0.6716 & 0.6462 \\
        YOLOv5  & 0.7244      & 0.6081    & 0.6612      & 0.6304         \\
        EfficientDet & 0.6939          &\textbf{0.6919} &\textbf{0.6929} & 0.6216   \\ \hline
      \hline
  \end{tabular}}
\end{table*}


Table \ref{table:Results} compares the performance of the benchmark algorithms in recall, precision, F1-Score and mAP. The Faster R-CNN networks achieved high recall, with the best recall method, Faster R-CNN Inception-v2-ResNet101, achieving the best result of 0.7554. However, the precision is worse than other networks due to the high number of false positives. On the other hand, EfficientDet has the best precision of 0.6919, which is comparable with its recall of 0.6939. When comparing the F1-Score, EfficientDet achieved the best result (0.6929), but had the poorest mAP (0.6216). The Faster R-CNN networks achieved better mAP, with the best result of 0.6596 with Faster R-CNN R-FCN. It should be noted that this paper aims to provide baseline results using existing end-to-end state-of-the-art object detection methods. Therefore, our experiments are based on the original settings of the networks, and do not include fine-tuning and post-processing. Better results may be achievable by using different anchor settings, for example, with YOLOv5, or by automated removal of duplicate detections.

To further analyse the results, Table \ref{tab:IoU} compares the performance of the networks on different IoU thresholds, from 0.50 to 0.90 with 0.10 increment.

\begin{table*}[]
	\centering
	\small\addtolength{\tabcolsep}{2pt}
	\renewcommand{\arraystretch}{1.5}
	\caption{Comparative performance of different networks for DFU detection on different IoU thresholds.}
	\label{tab:IoU}
	\scalebox{0.9}{
		\begin{tabular}{|c|c|c|c|c|c|c|c|c|c|c|c|c|}
			\cline{1-11}
			\multirow{2}{*}{Method} & 
			\multicolumn{2}{c|}{IoU$\geq$0.5}& 
			\multicolumn{2}{c|}{IoU$\geq$0.6}& 
			\multicolumn{2}{c|}{IoU$\geq$0.7}&
			\multicolumn{2}{c|}{IoU$\geq$0.8}&
			\multicolumn{2}{c|}{IoU$\geq$0.9}                             \\ \cline{2-11}
			& F1 & mAP & F1 & mAP & F1  & mAP & F1 & mAP  & F1 & mAP \\ \cline{1-11}
			R-FCN  &0.6784&0.6596& 0.6044 &0.5618 &0.4829 &0.4044  &0.2705  &0.1487  & 0.0534 &0.009\\ \cline{1-11}
			ResNet  &0.6623&0.6518  & 0.5931 & 0.5661 & 0.4701& 0.4087& 0.2703 &0.1689  &0.0551  &0.0112\\ \cline{1-11}
			Inc-Res  & 0.6716 & 0.6462 & 0.5902 &0.5385  &0.4592 &0.3827 &0.2616  &0.1644  &0.0483  &0.0095\\ \cline{1-11}
			YOLOv5  &0.6612 & 0.6304   &0.5898  &0.5353  &0.4418 &0.3420 &0.2355  &0.1175  &0.0383  & 0.0046 \\ \cline{1-11}
			EffDet  &0.6929 & 0.6216 & 0.6076 & 0.5143 & 0.4710&0.3503 &0.2505  &0.2167  &0.0343  &0.0031 \\ \cline{1-11}
	\end{tabular}}
\end{table*}




\section{Discussion}
\begin{figure}
	\centering
	\begin{tabular}{cc}
		\includegraphics[width=6cm,height=4.5cm]{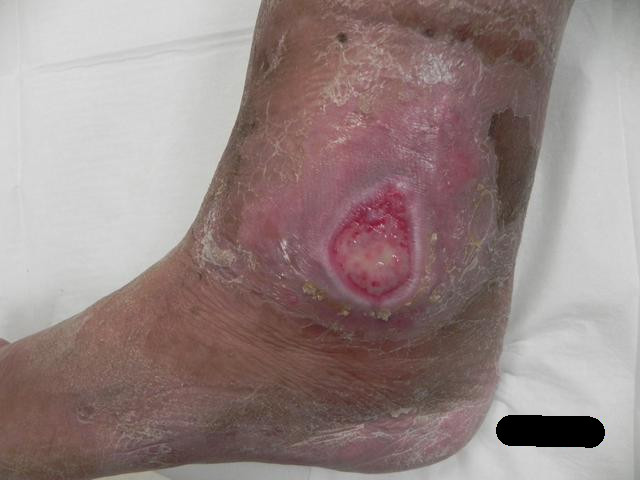} &
		\includegraphics[width=6cm,height=4.5cm]{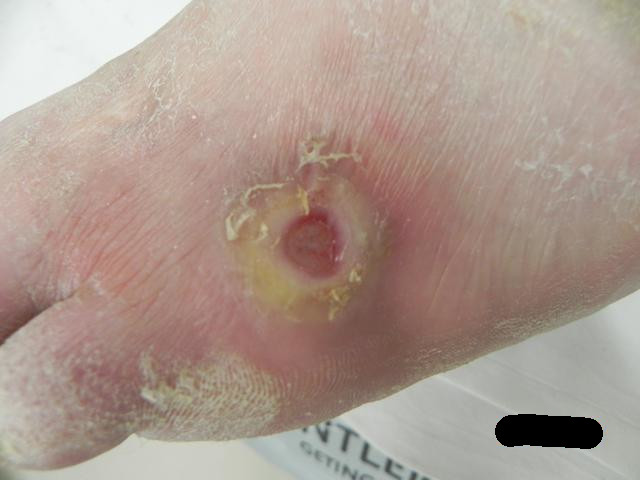}\\
		(a) &(b) \\  
	\end{tabular}     
	
	\caption[]{Easy cases where wounds are visible and detected by all networks.}
	\label{fig:easycases}
\end{figure}

\begin{figure}
	\centering
	\begin{tabular}{cc}
		\includegraphics[width=6cm,height=4.5cm]{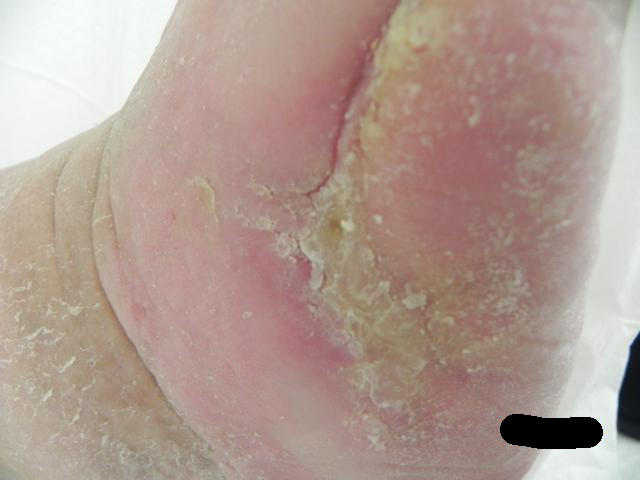} &
		\includegraphics[width=6cm,height=4.5cm]{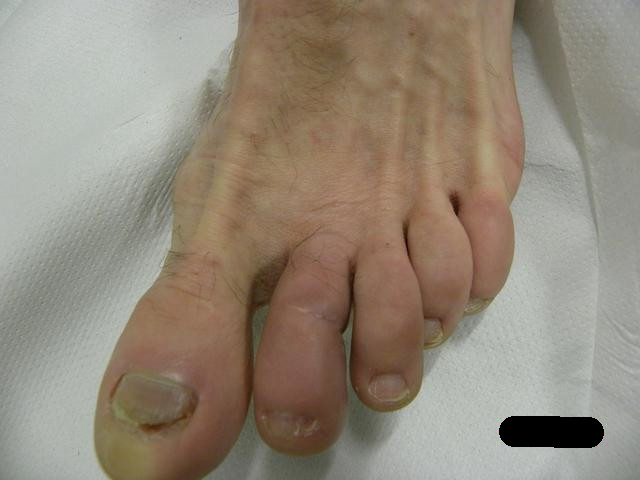}\\
		(a) &(b) \\  
	\end{tabular}     
	
	\caption[]{Difficult cases where wounds are not visible and not detected by all networks.}
	\label{fig:difficultcases}
\end{figure}

This paper provides baseline results for three popular deep learning object detection networks without manual pre-processing, fine-tuning or post-processing. We observe that these networks achieved comparable results. While others achieved better recall @0.5 IoU, EfficientDet shows a better trade-off between recall and precision, which yields the best F1-Score. In general, Faster R-CNN networks achieved better mAP@0.5 and remain the leader for mAP@0.6 to mAP@0.9 as shown in Table \ref{tab:IoU}. It is also noted that the F1-Score for Faster R-CNN is better than EfficientDet from @0.7 onwards. Figure \ref{fig:easycases} shows two easy cases detected by all networks, while Figure \ref{fig:difficultcases} shows two difficult cases that were missed by all networks.

Non-DFU images were included in our testing dataset to challenge the ability of each network. These images show various skin conditions on different regions of the body (not just the foot) and comprised of conditions such as melanoma, blisters, burns, gangrene, moles, insect bites, tinea pedis (athlete's foot), keloids, capillary hemangioma, shingles, chickenpox, chilblains, frostbite, onychomycosis (fungal toenail), ingrown toenail, intertrigo, mycosis fungoides, eczema, psoriasis. and ringworm. Many of these conditions share common visual traits with DFU. For the development of future networks, we will include images of these non-DFU conditions into a second classifier so that the model is more able to discern between DFU and non-DFU. The EfficientDet network trained in our experiments used the EfficientNet-B0 backbone, the smallest of the available EfficientNet backbones. Future work will assess the efficacy of the other available backbones (EfficientNet-B1 to B7) on our dataset. We will also investigate the ability of Generative Adversarial Networks (GANs) to generate convincing images of DFUs that could be used as data augmentation for future trained networks to improve on metrics such as mAP, Sensitivity and Specificity. Our initial experiments in this area have shown positive results.

\section{Conclusion}
This paper presents the largest DFU dataset made publicly available for the research community. The dataset provides an overview of the DFUC 2020 challenge held in conjunction with the MICCAI 2020 conference and reports the baseline results for the DFU test set using state-of-the-art object detection algorithms.  We will continue to make the dataset available for research after the challenge to motivate algorithm development in this domain. Additionally, we will report the results of the challenge in the near future. For our longer term plan, we will continue to collect and annotate DFU image data. It is expected that we will have more than 6000 images by early 2021, which will be made available for the DFUC 2021 Challenge \cite{yap2020dfuc2021}. This number is expected to then grow to 11,000 images for the DFUC 2022 Challenge. 

\section*{Acknowledgement}
We gratefully acknowledge the support of NVIDIA Corporation who provided access to GPU resources and sponsorship for DFUC 2020.

\addtocmark[2]{Author Index} 
\renewcommand{\indexname}{Author Index}
\printindex

\bibliographystyle{unsrt}
\bibliography{Ref}

\end{document}